\documentclass[11pt,a4paper]{article}
\usepackage[hyperref]{acl2019}
\usepackage{times}
\usepackage{latexsym}

\usepackage{url}

\usepackage{array}
\usepackage{color}

\usepackage{graphicx}
\usepackage[font=normalsize,position=bottom]{subfig}
\usepackage[noend]{algpseudocode}
\usepackage{multirow}

\usepackage{amssymb}
\usepackage{bm}
\usepackage{balance}
\usepackage{arydshln}
\usepackage{mathtools}
\usepackage{soul}
\usepackage[ruled]{algorithm2e}

\usepackage{booktabs}
\usepackage{amsmath,amsfonts}
\usepackage{xspace}

\usepackage{diagbox}

\usepackage{bbm}

\usepackage{xspace}
\usepackage{textcomp}

\usepackage{tabularx}
\usepackage{CJKutf8}
\usepackage[toc,page]{appendix}

\def\argmax{\mathop{\rm argmax}}%

\newcommand{\eg}{\emph{e.g. }} 
\newcommand{\ie}{\emph{i.e. }} 
\newcommand{\etc}{\emph{etc}}

\aclfinalcopy 
\def\aclpaperid{799} 

\title{Robust Neural Machine Translation with Doubly Adversarial Inputs}

\author{Yong Cheng, Lu Jiang, and Wolfgang Macherey\\
Google AI \\
  \texttt{\{chengyong, lujiang, wmach\}@google.com} \\
}

\date{}

\begin{document}
\maketitle
\begin{abstract}
Neural machine translation (NMT) often suffers from the vulnerability to noisy perturbations in the input. We propose an approach to improving the robustness of NMT models, which consists of two parts: (1) attack the translation model with adversarial source examples; 
(2) defend the translation model with adversarial target inputs to improve its robustness against the adversarial source inputs.
For the generation of adversarial inputs, we propose a gradient-based method to craft adversarial examples informed by the translation loss over the clean inputs.
Experimental results on Chinese-English and English-German translation tasks demonstrate that our approach achieves significant improvements ($2.8$ and $1.6$ BLEU points) over Transformer on standard clean benchmarks as well as exhibiting higher robustness on noisy data.
\end{abstract}

\section{Introduction}

In recent years, neural machine translation (NMT) has achieved tremendous success in advancing the quality of machine translation \cite{Wu:16,Hieber:17}. As an end-to-end sequence learning framework, NMT consists of two important components, the encoder and decoder, which are usually built on similar neural networks of different types, such as recurrent neural networks \cite{Sutskever:14,Bahdanau:15,Chen:18}, convolutional neural networks \cite{Gehring:17}, and more recently on transformer networks \cite{Vaswani:17}. 
To overcome the bottleneck of encoding the entire input sentence into a single vector, an attention mechanism was introduced, which further enhanced translation performance \cite{Bahdanau:15}.
Deeper neural networks with increased model capacities in NMT have also been explored and shown promising results \cite{Bapna:18}.

Despite these successes, NMT models are still vulnerable to perturbations in the input sentences.
For example, \citet{Belinkov:17} found that NMT models can be immensely brittle to small perturbations applied to the inputs. Even if these perturbations are not strong enough to alter the meaning of an input sentence, they can nevertheless result in different and often incorrect translations.
Consider the example in Table~\ref{table:wrong_translation}, the Transformer model will generate a worse translation (revealing gender bias) for a minor change in the input from ``he'' to ``she''.
Perturbations originate from two sources: (a) natural noise in the annotation and (b) artificial deviations generated by attack models. In this paper, we do not distinguish the source of a perturbation and term perturbed examples as adversarial examples. The presence of such adversarial examples can lead to significant degradation of the generalization performance of the NMT model. 

\begin{table}
\begin{tabular}{c|l}
\hline
 Input & \begin{CJK*}{UTF8}{gbsn}{\small \textcolor{red}{他}(\textcolor{red}{她})一个残疾人，我女儿身体好好地。}\end{CJK*}\\
        \hline
 
 Original  & \textbf{he is a} handicapped \textbf{person}, my \\
 Output &daughter is in good health. $\checkmark$\\
 \hline
 Perturbed    &\textbf{one of her} handicapped \textbf{people}, my\\
 Output      &daughter is in good health.  $\mathbf{\times}$\\
\hline
\end{tabular}
\caption{An example of Transformer NMT translation result for an input and its perturbed input by replacing ``\begin{CJK*}{UTF8}{gbsn}{他}\end{CJK*}(he)'' to ``\begin{CJK*}{UTF8}{gbsn}{她}\end{CJK*}(she)''.}
\label{table:wrong_translation} 
\end{table}

A few studies have been proposed in other natural language processing (NLP) tasks aiming to tackle this issue in classification tasks, \eg in~\cite{Miyato:17,Alzantot:18,Ebrahimi:18,Zhao:18}. As for NMT, previous approaches relied on prior knowledge to generate adversarial examples to improve the robustness, neglecting specific downstream NMT models. For example, \citet{Belinkov:17} and \citet{Karpukhin:19} studied how to use some synthetic noise and/or natural noise. \citet{Cheng:18} proposed adversarial stability training to improve the robustness on arbitrary noise type including feature-level and word-level noise. \citet{Liu:18} examined the homophonic noise for Chinese translation.

This paper studies learning a robust NMT model that is able to overcome small perturbations in the input sentences. Different from prior work, our work deals with the perturbed examples jointly generated by a white-box NMT model, which means that we have access to the parameters of the attacked model. To the best of our knowledge, the only previous work on this topic is from~\cite{Ebrahimi:18b} on character-level NMT.
Overcoming adversarial examples in NMT is a challenging problem as the words in the input are represented as discrete variables, making them difficult to be switched by imperceptible perturbations. Moreover, the characteristics of sequence generation in NMT further intensify this difficulty.
To tackle this problem, we propose a gradient-based method, $AdvGen$, to construct adversarial examples guided by the final translation loss from the clean inputs of a NMT model. $AdvGen$ is applied to both encoding and decoding stages: (1) we attack a NMT model by generating adversarial source inputs that are sensitive to the training loss; (2) we then defend the NMT model with the adversarial target inputs, aiming at reducing the prediction errors for the corresponding adversarial source inputs.

Our contribution is threefold:
\begin{enumerate}
    \item A white-box method to generate adversarial examples is explored for NMT. Our method is a gradient-based approach guided by the translation loss.
    \item We propose a new approach to improving the robustness of NMT with doubly adversarial inputs. The adversarial inputs in the encoder aim at attacking the NMT models, while those in the decoder are capable of defending the errors in predictions.
    \item Our approach achieves significant improvements over the previous state-of-the-art Transformer model on two common translation benchmarks. 
\end{enumerate}

Experimental results on the standard Chinese-English and English-German translation benchmarks show that our approach yields an improvement of $2.8$ and $1.6$ BLEU points over the state-of-the-art models including Transformer~\cite{Vaswani:17}. This result substantiates that our model improves the generalization performance over the clean benchmark datasets. Further experiments on noisy text verify the ability of our approach to improving robustness. We also conduct ablation studies to gain further insight into which parts of our approach matter the most.

\section{Background}
\noindent\textbf{Neural Machine Translation}
NMT is typically a neural network with an encoder-decoder architecture. It aims to maximize the likelihood of a parallel corpus $\mathcal{S} = \{( \mathbf{x}^{(s)}, \mathbf{y}^{(s)}) \}_{s=1}^{|\mathcal{S}|}$. Different variants derived from this architecture have been proposed recently \cite{Bahdanau:15, Gehring:17, Vaswani:17}. This paper focuses on the recent Transformer model \cite{Vaswani:17} due to its superior performance, although our approach seems applicable to other models, too.

The encoder in NMT maps a source sentence $\mathbf{x} = x_{1},...,x_{I}$ to a sequence of $I$ word embeddings $e(\mathbf{x}) = e(x_{1}),...,e(x_{I})$. 
Then the word embeddings are encoded to their corresponding continuous hidden representations $\mathbf{h}$ by the transformation layer. 
Similarly, the decoder maps its target input sentence $\mathbf{z} = z_{1},...,z_{J}$ to a sequence of $J$ word embeddings. For clarity, we denote the input and output in the decoder as $\mathbf{z}$ and $\mathbf{y}$. $\mathbf{z}$ is a shifted copy of $\mathbf{y}$ in the standard NMT model, \ie $\mathbf{z} = \langle sos \rangle, \mathbf{y}_{1}, \cdots, \mathbf{y}_{J-1}$, where $\langle sos \rangle$ is a start symbol. Conditioned on the hidden representations $\mathbf{h}$ and the target input $\mathbf{z}$, the decoder generates $\mathbf{y}$ as:
\begin{eqnarray}
P(\mathbf{y}|\mathbf{x};\bm{\theta}_{mt}) = \prod_{j=1}^{J} P(y_{j}|\mathbf{z}_{\leq j}, \mathbf{h};\bm{\theta}_{mt})
\end{eqnarray} 
where $\bm{\theta}_{mt}$ is a set of model parameters and $\mathbf{z}_{<j}$ is a partial target input. The training loss on $\mathcal{S}$ is defined as:
\begin{eqnarray}
\mathcal{L}_{clean}(\bm{\theta}_{mt}) = \frac{1}{|S|} \sum_{( \mathbf{x}, \mathbf{y} ) \in S} -\log P(\mathbf{y}|\mathbf{x};\bm{\theta}_{mt})
\end{eqnarray}

\noindent\textbf{Adversarial Examples Generation}
An adversarial example is usually constructed by corrupting the original input with a small perturbation such that the difference to the original input remains less perceptible but dramatically distorts the model output. The adversarial examples can be generated by a white-box or black-box model, where the latter does not have access to the attacked models and often relies on prior knowledge. The former white-box examples are generated using the information of the attacked models. Formally, a set of adversarial examples $\mathcal{Z}(\mathbf{x}, y)$ is generated with respect to a training sample $(\mathbf{x}, y)$ by solving an optimization problem:
\begin{eqnarray}
\left\{\mathbf{x}^{\prime} \!\mid\! \mathcal{R}(\mathbf{x}^{\prime}, \mathbf{x}) \!\leq\! \epsilon, \argmax_{\mathbf{x}^{\prime}} J(\mathbf{x}^{\prime}, y; \bm{\theta}) \right\} \label{eq:adv_set}
\end{eqnarray}
where $J(\cdot)$ measures the possibility of a sample being adversarial, and $\mathcal{R}(\mathbf{x}^{\prime}, \mathbf{x})$ captures the degree of imperceptibility for a perturbation. For example, in the classification task, $J(\cdot)$ is a function outputting the most possible target class $y^{\prime}$ ($y^{\prime} \neq y$) when fed with the adversarial example $\mathbf{x}^{\prime}$. Although it is difficult to give a precise definition of the degree of imperceptibility $\mathcal{R}(\mathbf{x}^{\prime}, \mathbf{x})$, $l_{\infty}$ norm is usually used to bound the perturbations in image classification \cite{Goodfellow:14b}.


\section{Approach}
Our goal is to learn robust NMT models that can overcome small perturbations in the input sentences. 
As opposed to images, where small perturbations to pixels are imperceptible, even a single word change in natural languages can be perceived. NMT is a sequence generation model wherein each output word is conditioned on all previous predictions. Thus, one question is how to design meaningful perturbation operations for NMT.

We propose a gradient-based approach, called $AdvGen$, to construct adversarial examples and use these examples to both attack as well as defend the NMT model. Our intuition is that an ideal model would generate similar translation results for similar input sentences despite any small difference caused by perturbations.

The attack and defense are carried out in the end-to-end training of the NMT model.
We first use $AdvGen$ to construct an adversarial example $\mathbf{x}^{\prime}$ from the original input $\mathbf{x}$ to attack the NMT model. We then use $AdvGen$ to find an adversarial target input $\mathbf{z}^{\prime}$ from the decoder input $\mathbf{z}$ to improve the NMT model robustness to adversarial perturbations in the source input $\mathbf{x}^{\prime}$. Thereby we hope the NMT model will be robust against both the source adversarial input $\mathbf{x}^{\prime}$ and adversarial perturbations in target predictions $\mathbf{z}^{\prime}$.
The rest of this section will discuss the attack and defense procedures in detail.

\subsection{Attack with Adversarial Source Inputs}\label{sec:attack_source}
Following~\cite{Goodfellow:14b,Miyato:17,Ebrahimi:18}, we study the white-box method to generate adversarial examples tightly guided by the training loss. Given a parallel sentence pair $(\mathbf{x},\mathbf{y})$, according to Eq.~\eqref{eq:adv_set}, we generate a set of adversarial examples $\mathcal{A}(\mathbf{x}, \mathbf{y})$ specific to the NMT model by:
\begin{eqnarray}
\left\{ \mathbf{x}^{\prime} \!\mid \!\mathcal{R}(\mathbf{x}^{\prime}, \mathbf{x}) \! \leq \! \epsilon, \argmax_{\mathbf{x}^{\prime}} -\!\log P(\mathbf{y}|\mathbf{x}^{\prime}; \bm{\theta}_{mt}) \right\} \label{eq:nmtadv_set}
\end{eqnarray}
where we use the negative log translation probability $-\log P(\mathbf{y}|\mathbf{x}^{\prime};\bm{\theta}_{mt})$ to estimate $J(\cdot)$ in Eq.~\eqref{eq:adv_set}. 
The formula constructs adversarial examples that are expected to distort the current prediction and retain semantic similarity bounded by $\mathcal{R}$.

It is intractable to obtain an exact solution for Eq.~\eqref{eq:nmtadv_set}. 
We therefore resort to a greedy approach based on the gradient to circumvent it.
For the original input $\mathbf{x}$, we induce a possible adversarial word $x_{i}^{\prime}$ for the word $x_{i}$ in $\mathbf{x}$: 
\begin{eqnarray}
x_{i}^{\prime} &=& \argmax_{x \in \mathcal{V}_{x}} \mathop{\rm sim} \left( e(x) - e(x_{i}), \mathbf{g}_{x_{i}} \right) \label{eq:select_adv} \\
\mathbf{g}_{x_{i}} &=&  \nabla_{e(x_{i})} - \log P(\mathbf{y}|\mathbf{x}; \bm{\theta}) \label{eq:adv_gradient}
\end{eqnarray}
where $\mathbf{g}_{x_{i}}$ is a gradient vector wrt.~$e(x_{i})$, $\mathcal{V}_{x}$ is the vocabulary for the source language, and $\mathop{\rm sim}(\cdot, \cdot)$ denotes the similarity function by calculating the cosine distance between two vectors.

Eq.~\eqref{eq:select_adv} enumerates all words in $\mathcal{V}_{x}$ incurring formidable computational cost. We hence substitute it with a dynamic set $\mathcal{V}_{x_{i}}$ that is specific for each word $x_{i}$. Let $Q(x_i, \mathbf{x}) \in \mathbb{R}^{|\mathcal{V}|}$ denote the likelihood of the $i$-th word in the sentence $\mathbf{x}$. Define $\mathcal{V}_{x_i} = top\_n (Q(x_i, \mathbf{x}))$ as the set of the $n$ most probable words among the top $n$ scores in terms of $Q(x_i, \mathbf{x})$, where $n$ is a small constant integer and $\vert \mathcal{V}_{x_{i}} \vert  \ll \vert \mathcal{V}_{x} \vert$.
For the source, we estimate it from:
\begin{eqnarray}
Q_{src}(x_i, \mathbf{x}) = P_{lm}(x|\mathbf{x}_{< i}, \mathbf{x}_{> i}; \bm{\theta}_{lm}^{x})
\label{eq:q_src}
\end{eqnarray}
Here, $P_{lm}$ is a bidirectional language model for the source language.


The introduction of language model has three benefits.
First, it enables a computationally feasible way to approximate Eq.~\eqref{eq:select_adv}. Second, the language model can retain the semantic similarity between the original words and their adversarial counterparts to strengthen the constraint $\mathcal{R}$ in Eq.~\eqref{eq:nmtadv_set}. Finally, it prevents word representations from being degenerative because replacements with adversarial words usually affect the context information around them.

Algorithm \ref{algo1} describes the function $AdvGen$ for generating an adversarial sentence  $\mathbf{s}^{\prime}$ from an input sentence $\mathbf{s}$. 
The function inputs are: $Q$ is a likelihood function for the candidate set generation, and for the source, it is $Q_{src}$ from Eq.~\eqref{eq:q_src}. $D_{pos}$ is a distribution over the word position $\{1,..,\vert \mathbf{x} \vert \}$ from which the adversarial word is sampled. For the source, we use the simple uniform distribution $\mathcal{U}$.
Following the constraint $\mathcal{R}$, we want the output sentence not to deviate too much from the input sentence and thus only change a small fraction of its constituent words based on a hyper-parameter $\gamma \in [0,1]$.



\begin{algorithm}[!t]
\SetAlgoLined
\LinesNumbered
\KwIn{$\mathbf{s}$: Input sentence, $Q$: Likelihood function,
	  $\mathcal{D}_{pos}$: Distribution for word sampling,
	  $\mathcal{L}$: translation loss.}
\KwOut{$\mathbf{s}^{\prime}$: Output adversarial sentence}
\SetKwFunction{algo}{AdvGen}
\SetKwProg{Fn}{Function}{:}{}
\Fn{\algo{$\mathbf{s}$, $Q$, $D_{pos}$, $\mathcal{L}$}}{
$ POS  \leftarrow$ sample $\gamma \vert \mathbf{s} \vert$ positions from $\{1,...,\vert \mathbf{s} \vert \}$ according to $D_{pos}$  {\footnotesize\tcp{$\gamma$ is a sampling ratio}}

 \ForEach{$i \in \{1,...,\vert \mathbf{s} \vert \}$ }{
  \eIf{$i \in POS$}{
  	$\mathcal{V}_{s_{i}} \leftarrow top\_n (Q(s_{i}, \mathbf{s}))-\{ s_{i}\}$\;
  	
	$\mathbf{g}_{s_{i}} \leftarrow  \nabla_{e(s_{i})} \mathcal{L}$\;
	

	Compute $s_{i}^{\prime}$ by Eq.~\eqref{eq:select_adv}\;
   }{
   	$s_{i}^{\prime} \leftarrow s_{i}$\;
  }
 }
 \KwRet $\mathbf{s}^{\prime}$ }
 \caption{The $AdvGen$ Function.} \label{algo1}
\end{algorithm}

\subsection{Defense with Adversarial Target Inputs}
After generating an adversarial example $\mathbf{x}^{\prime}$, we treat $(\mathbf{x}^{\prime}, \mathbf{y})$ as a new training data point to improve the model's robustness. These adversarial examples in the source tend to introduce errors which may accumulate and cause drastic changes to the decoder prediction. To defend the model from errors in the decoder predictions, 
we generate an adversarial target input by $AdvGen$, similar to what we discussed in Section~\ref{sec:attack_source}. The decoder trained with the adversarial target input is expected to be more robust to the small perturbations introduced in the source input. The ablation study results in Table~\ref{table:ablation_study} substantiate the benefit of this defense mechanism.


Formally, let $\mathbf{z}$ be the decoder input for the sentence pair $(\mathbf{x}, \mathbf{y})$. We use the same $AdvGen$ function to generate an adversarial target input $\mathbf{z}^{\prime}$ from $\mathbf{z}$ by:
\begin{eqnarray}
\mathbf{z}^{\prime} = 
AdvGen(\mathbf{z}, Q_{trg}, D_{trg}, -\log P(\mathbf{y}|\mathbf{x}^{\prime}))
\end{eqnarray}

Note that for the target, the translation loss in Eq.~\eqref{eq:adv_gradient} is replaced by $-\log P(\mathbf{y}|\mathbf{x}^{\prime})$. $Q_{trg}$ is the likelihood for selecting the target word candidate set $\mathcal{V}_{z}$.
To compute it, we combine the NMT model prediction with a language model $P_{lm}(\mathbf{y};\bm{\theta}_{lm}^{y})$ as follow:
\begin{eqnarray}
\label{eq:trg_can}
Q_{trg}(z_i, \mathbf{z}) = &
 \lambda P(z|\mathbf{z}_{<i}, \mathbf{z}_{>i};\bm{\theta}_{lm}^{y})   \quad \quad \quad \nonumber \\
 &+ (1-\lambda)P(z| \mathbf{z}_{<i}, \mathbf{x}^{\prime}; \bm{\theta}_{mt}) 
\end{eqnarray}
where $\lambda$ balances the importance between two models. 

$D_{trg}$ is a distribution for sampling positions for the target input. Different from the uniform distribution used in the source, in the target sentence we want to change those relevant words influenced by the perturbed words in the source input.
To do so, we use the attention matrix $\mathcal{M}$ learned in the NMT model, obtained at the current mini-batch, to compute the distribution over $(\mathbf{x}, \mathbf{y}, \mathbf{x}^{\prime})$ by:
\begin{eqnarray}
P(j) = \frac{\sum_{i} \mathcal{M}_{ij} \delta(x_{i}, x^{\prime}_{i})}{\sum_{k}\sum_{i} \mathcal{M}_{ik} \delta(x_{i}, x^{\prime}_{i})}, j \in \{1,..,\vert \mathbf{y} \vert \} \label{eq:trg_pos}
\end{eqnarray}
where $\mathcal{M}_{ij}$ is the attention score between $x_{i}$ and $y_{j}$ and $\delta(x_{i},x_{i}^{\prime})$ is an indicator function that yields $1$ if $x_{i} \neq x_{i}^{\prime}$ and $0$ otherwise.


\begin{algorithm}[!t]
\SetAlgoLined
\LinesNumbered
\KwIn{$(\mathbf{x},\mathbf{y})$: a parallel sentence pair}
\KwOut{$loss$: a robustness loss for $(\mathbf{x}, \mathbf{y})$}
\SetKwFunction{algo}{RobustLoss}
\SetKwProg{Fn}{Function}{:}{}
\Fn{\algo{$\mathbf{x}, \mathbf{y}$}}{

    Initialize the sampling ratio $\gamma_{src}$ and $\gamma_{trg}$\;

	Compute $Q_{src}$ by Eq.~\eqref{eq:q_src}\;
	
	Set $D_{src}$ as a uniform distribution\;
	
	{\small $\mathbf{x}^{\prime} \leftarrow AdvGen(\mathbf{x}, Q_{src}, D_{src}, -\log P(\mathbf{y}|\mathbf{x}))$\;}
	
	$Q_{trg}$ is computed as Eq.~\eqref{eq:trg_can}\;
	
	$D_{trg}$ is computed as Eq.~\eqref{eq:trg_pos}\;
	
	{\small $\mathbf{z}^{\prime} \leftarrow AdvGen(\mathbf{z}, Q_{trg}, D_{trg}, -\log P(\mathbf{y}|\mathbf{x}^{\prime}))$\;}
	
	$loss \leftarrow -\log P(\mathbf{y}|\mathbf{x}^{\prime}, \mathbf{z}^{\prime}; \bm{\theta}_{mt})$ 
	
 \KwRet $loss$ }
 \caption{Computing Robustness Loss.} \label{algo2}
\end{algorithm}

\begin{table*}[!t]
\centering
\begin{tabular}{l|l||l|lllll}
\hline
{Method} &{Model} &{MT06} &{MT02} & {MT03} & {MT04} & {MT05} &{MT08}\\
\hline \hline
\newcite{Vaswani:17} &Trans.-Base &44.59 &44.82 &43.68 &45.60 &44.57 &35.07 \\
\hline
\newcite{Miyato:17} &Trans.-Base &45.11 &45.95 &44.68 &45.99 &45.32 &35.84 \\
\hline
\newcite{Sennrich:16c} &Trans.-Base &44.96 &46.03 &44.81 &46.01 &45.69  &35.32\\
\hline
\newcite{Wang:18} &Trans.-Base &45.47 &46.31 &45.30 &46.45 &45.62 &35.66 \\
\hline
\multirow{2}{*}{\newcite{Cheng:18}} &RNMT$_{lex.}$ &43.57 &44.82 &42.95 &45.05 &43.45 &34.85 \\
&RNMT$_{feat.}$ &44.44 &46.10 &44.07 &45.61 &44.06 &34.94 \\
\hline
\multirow{2}{*}{\newcite{Cheng:18}}
&Trans.-Base$_{feat.}$ &45.37 &46.16 &44.41 &46.32 &45.30 &35.85\\
&Trans.-Base$_{lex.}$ &45.78 &45.96 &45.51 &46.49 &45.73 &36.08 \\
\hline

\newcite{Sennrich:16b}* &Trans.-Base &46.39 &47.31 &47.10 &47.81 &45.69  &36.43\\
\hline \hline
Ours &Trans.-Base &46.95 &47.06 &46.48 &47.39 &46.58 &37.38 \\
Ours + BackTranslation* &Trans.-Base &\textbf{47.74} &\textbf{48.13} &\textbf{47.83} &\textbf{49.13} &\textbf{49.04} &\textbf{38.61} \\
\hline

\end{tabular}
\caption{Comparison with baseline methods trained on different backbone models (second column). * indicates the method trained using an extra corpus.} \label{table:comparison}
\end{table*}

\begin{table*}[!t]
\centering
\begin{tabular}{l|l||l|lllll}
\hline
{Method} &{Model} &{MT06} &{MT02} & {MT03} & {MT04} & {MT05} &{MT08}\\
\hline \hline  

\newcite{Vaswani:17} &Trans.-Base &44.59 &44.82 &43.68 &45.60 &44.57 &35.07 \\
\hline \hline
Ours &Trans.-Base &\textbf{46.95} &\textbf{47.06} &\textbf{46.48} &\textbf{47.39} &\textbf{46.58} &\textbf{37.38} \\
\hline
\end{tabular}
\caption{Results on NIST Chinese-English translation.} \label{table:zh-en}
\label{table:comparison_zhen}
\end{table*}

\begin{table}[!t]
\centering
\begin{tabular}{l|l||l}
\hline
Method &Model  &BLEU \\
\hline \hline
\multirow{2}{*}{\citeauthor{Vaswani:17}} & Trans.-Base  & 27.30 \\
& Trans.-Big  & 28.40 \\
\hline
\citeauthor{Chen:18} & RNMT+ & 28.49 \\
\hline \hline
\multirow{2}{*}{Ours} & Trans.-Base  & 28.34 \\
& Trans.-Big  &\textbf{30.01} \\
\hline
\end{tabular}
\caption{Results on WMT'14 English-German translation.} \label{table:en-de}
\label{table:comparison_ende}
\end{table}

\subsection{Training}
Algorithm \ref{algo2} details the entire procedure to calculate the robustness loss for a parallel sentence pair $( \mathbf{x}, \mathbf{y} )$.
We run $AdvGen$ twice to obtain $\mathbf{x}^{\prime}$ and $\mathbf{z}^{\prime}$. We do not backpropagate gradients over $AdvGen$ when updating parameters, which just plays a role of data generator. In our implementation, this function incurs at most a $20\%$ time overhead compared to the standard Transformer model. Accordingly, we compute the robustness loss on $\mathcal{S}$ as:
\begin{eqnarray}
\mathcal{L}_{robust}(\bm{\theta}_{mt}) \!=\! \frac{1}{\vert \mathcal{S} \vert} \!\!\sum_{( \mathbf{x}, \mathbf{y} ) \in \mathcal{S}} \!\!\!-\!\log P(\mathbf{y}|\mathbf{x}^{\prime}, \mathbf{z}^{\prime}; \bm{\theta}_{mt})
\label{eq:robust_loss}
\end{eqnarray}

The final training objective $\mathcal{L}$ is a combination of four loss functions:
\begin{eqnarray}
\mathcal{L}(\bm{\theta}_{mt}, \bm{\theta}_{lm}^{x}, \bm{\theta}_{lm}^{y}) = \mathcal{L}_{clean}(\bm{\theta}_{mt}) + \mathcal{L}_{lm}(\bm{\theta}_{lm}^{x})   \nonumber \\
+ \mathcal{L}_{robust}(\bm{\theta}_{mt}) + \mathcal{L}_{lm}(\bm{\theta}_{lm}^{y})
\end{eqnarray}
where $\bm{\theta}_{lm}^{x}$ and $\bm{\theta}_{lm}^{y}$ are two sets of model parameters for source and target bidirectional language models, respectively. The word embeddings are shared between $\bm{\theta}_{mt}$ and $\bm{\theta}_{lm}^{x}$ and likewise between $\bm{\theta}_{mt}$ and $\bm{\theta}_{lm}^{y}$.


\section{Experiments}

\subsection{Setup}
We conducted experiments on Chinese-English and English-German translation tasks. The Chinese-English training set is from the LDC corpus that compromises 1.2M sentence pairs. We used the NIST 2006 dataset as the validation set for model selection and hyper-parameters tuning, and NIST 2002, 2003, 2004, 2005, 2008 as test sets. For the English-German translation task, we used the WMT'14 corpus consisting of 4.5M sentence pairs. The validation set is newstest2013, and the test set is newstest2014.

In both translation tasks, we merged the source and target training sets and used byte pair encoding (BPE) \cite{Sennrich:16a} to encode words through sub-word units. 
We built a shared vocabulary of 32K sub-words for English-German and created shared BPE codes with 60K operations for Chinese-English that induce two vocabularies with 46K Chinese sub-words and 30K English sub-words. 
We report case-sensitive tokenized BLEU scores for English-German and case-insensitive tokenized BLEU scores for Chinese-English \cite{Papineni:02}. For a fair comparison, we did not average multiple checkpoints \cite{Vaswani:17}, and only report results on a single converged model.

We implemented our approach based on the Transformer model \cite{Vaswani:17}. In $AdvGen$, We modified multiple positions in the source and target input sentences in parallel.
The bidirectional language model used in $AdvGen$ consists of left-to-right and right-to-left Transformer networks, a linear layer to combine final representations from these two networks, and a softmax layer to make predictions. The Transformer network was built using six transformation layers which keeps consistent with the encoder in the Transformer model. 
The hyperparameters in the Transformer model were set according to the default values described in \cite{Vaswani:17}. 
We denote the Transformer model with $512$ hidden units as Trans.-Base and $1024$ hidden units as Trans.-Big.

We tuned the hyperparameters in our approach on the validation set via a grid search. Specifically, $\lambda$ was set to $0.5$. The $n$ in $top\_n$ to select word candidates was set to $10$. 
The ratio pair $(\gamma_{src}, \gamma_{trg})$ was set to $(0.25, 0.50)$ with the exception of Trans.-Base on English-German where it was set to $(0.15, 0.15)$. We treated the single part of parallel corpus as monolingual data to train bidirectional language models without introducing additional data. The model parameters in our approach were trained from scratch except for the parameters in language models initialized by the models pre-trained on the single part of parallel corpus. The parameters of language models were still updated during robustness training.

\subsection{Main Results}
Table \ref{table:zh-en} shows the BLEU scores on the NIST Chinese-English translation task.
We first compare our approach with the Transformer model \cite{Vaswani:17} on which our model is built.
As we see, the introduction of our method to the standard backbone model (Trans.-Base) leads to substantial improvements across the validation and test sets. Specifically, our approach achieves an average gain of $2.25$ BLEU points and up to $2.8$ BLEU points on NIST03.

Table \ref{table:en-de} shows the results on WMT'14 English-German translation. We compare our approach with Transformer for different numbers of hidden units (\ie{1024 and 512}) and a related RNN-based NMT model RNMT+ \cite{Chen:18}. As is shown in Table ~\ref{table:en-de}, our approach achieves improvements over the Transformer for the same number of hidden units, \ie $1.04$ BLEU points over Trans.-Base, $1.61$ BLEU points over Trans.-Big, and $1.52$ BLEU points over RNMT+ model.
Recall that our approach is built on top of the Transformer model. The notable gain in terms of BLEU verifies our English-German translation model.

\begin{table*}[!t]
\centering

\begin{tabular}{c|l}
\hline
 Input \& Noisy Input & \begin{CJK*}{UTF8}{gbsn}{这体现了中俄两国和两国议会间{\color{red}{密切}}({\color{red}{紧密}})的友好合作关系。}\end{CJK*}\\
 \hline
 Reference & this expressed the relationship of close friendship and cooperation between \\
        & China and Russia and between our parliaments. \\
\hline \hline
\citeauthor{Vaswani:17} & this reflects the close friendship and cooperation between \textbf{China and Russia} \\
on Input       &and \textbf{between} the parliaments \textbf{of} the two countries. \\
\hline

\citeauthor{Vaswani:17} & this reflects the close friendship and cooperation between the two countries                        \\
on Noisy Input        &and the \textbf{two} parliaments. \\
\hline \hline
Ours & this \textbf{reflects} the close relations of friendship and cooperation between China \\
on Input    &and Russia and between their parliaments. \\
\hline
Ours & this \textbf{embodied} the close relations of friendship and cooperation between China   \\
on Noisy Input        &and Russia and between their parliaments.\\
\hline
\end{tabular}
\caption{Comparison of translation results of Transformer and our model for an input and its perturbed input.}
\label{table:robust_examples} 
\end{table*}

\begin{table}[!t]
\centering
\begin{tabular}{c||c|c|c|c}
\hline
Method &0.00 &0.05 &0.10 &0.15 \\
\hline \hline
\citeauthor{Vaswani:17} &44.59 &41.54 &38.84 &35.71\\
\citeauthor{Miyato:17} &45.11 &42.11 &39.39 &36.44\\
\citeauthor{Cheng:18} &45.78 &42.90 &40.58 &38.46\\
\hline \hline
Ours &\textbf{46.95} &\textbf{44.20} &\textbf{41.71} &\textbf{39.89} \\
\hline
\end{tabular}
\caption{Results on artificial noisy inputs. The column lists results for different noise fractions.}
\label{table:robust} 
\end{table}

\begin{table}[!t]
\centering
\begin{tabular}{c||c|c|c|c}
\hline
Method &0.00 &0.05 &0.10 &0.15 \\
\hline \hline
\citeauthor{Vaswani:17} &100 &77.08 &62.00 &52.50\\
\citeauthor{Miyato:17} &100 &79.19 &63.12 &53.51\\
\citeauthor{Cheng:18} &100 &79.66 &65.16 &56.11\\
\hline \hline
Ours &100 &\textbf{82.76} &\textbf{69.23} &\textbf{60.70} \\
\hline
\end{tabular}
\caption{BLEU scores computed using the zero noise fraction output as a reference.} \label{table:robust_change} 
\end{table}

\subsection{Comparison to Baseline Methods} 
To further verify our method, we compare to recent related techniques for robust NMT learning methods.
For a fair comparison, we implemented all methods on the same Transformer backbone.

\citet{Miyato:17} applied perturbations to word embeddings using adversarial learning in text classification tasks. We apply this method to the NMT model.

\citet{Sennrich:16c} augmented the training data with word dropout. We follow their method to randomly set source word embeddings to zero with the probability of $0.1$. This simple technique performs reasonably well on the Chinese-English translation.

\citet{Wang:18} introduced a data-augmentation method for NMT called SwitchOut to randomly replace words in both source and target sentences with other words. 

\citet{Cheng:18} employed adversarial stability training to improve the robustness of NMT. We cite their numbers reported in the paper for the RNN-based NMT backbone and implemented their method on the Transformer backbone. We consider two types of noisy perturbations in their method and use subscripts {\em lex.} and {\em fea.} to denote them.

\citet{Sennrich:16b} is a common data-augmentation method for NMT. The method back-translates monolingual data by an inverse translation model. We sampled 1.2M English sentences from the Xinhua portion of the GIGAWORD corpus as monolingual data.
We then back-translated them with an English-Chinese NMT model and re-trained the Chinese-English model using back-translated data as well as original parallel data.

Table \ref{table:comparison} shows the comparisons to the above five baseline methods.
Among all methods trained without extra corpora, our approach achieves the best result across datasets. After incorporating the back-translated corpus, our method yields an additional gain of 1-3 points over \cite{Sennrich:16b} trained on the same back-translated corpus. Since all methods are built on top of the same backbone, the result substantiates the efficacy of our method on the standard benchmarks that contain natural noise. Compared to \cite{Miyato:17}, we found that continuous gradient-based perturbations to word embeddings can be absorbed quickly, often resulting in a worse BLEU score than the proposed discrete perturbations by word replacement.


\subsection{Results on Noisy Data}
We have shown improvements on the standard clean benchmarks. This subsection validates the robustness of the NMT models over artificial noise. To this end, we added synthetic noise to the clean validation set by randomly replacing a word with a relevant word according to the similarity of their word embeddings. We repeated the process in a sentence according to a pre-defined noise fraction where a noise level of $0.0$ yields the original clean dataset while $1.0$ provides an entirely altered set. For each sentence, we generated $100$ noisy sentences. We then re-scored those sentences using a pre-trained bidirectional language model, and picked the best one as the noisy input.


Table \ref{table:robust} shows results on artificial noisy inputs. BLEU scores were computed against the ground-truth translation result. As we see, our approach outperforms all baseline methods across all noise levels. The improvement is generally more evident when the  noise fraction becomes larger.

To further analyze the prediction stability, we compared the model outputs for clean and noisy inputs.
To do so, we selected the output of a model on clean input (noise fraction equals 0.0) as a reference and computed the BLEU score against this reference. Table \ref{table:robust_change} presents the results where the second column 100 means that the output is exactly the same as the reference. The relative drop of our model, as the noise level grows, is smaller compared to other baseline methods. The results in Table~\ref{table:robust} and Table~\ref{table:robust_change} together suggest our model is more robust toward the input noise.


Table ~\ref{table:robust_examples} shows an example translation (More examples are shown in the Appendix). In this example, the original and noisy input have literally the same meaning, where ``\begin{CJK*}{UTF8}{gbsn}{密切}\end{CJK*}'' and ``\begin{CJK*}{UTF8}{gbsn}{紧密}\end{CJK*}'' both mean ``close'' in Chinese. Our model retains very important words such as ``China and Russia'', which are missing in the Transformer results.

\begin{table}[!t]
\centering
\begin{tabular}{c|c|c|c||c}
\hline
\multirow{2}{*}{$\mathcal{L}_{clean}$} & \multicolumn{2}{c|}{$\mathcal{L}_{robust}$} & \multirow{2}{*}{$\mathcal{L}_{lm}$} & \multirow{2}{*}{BLEU} \\
\cline{2-3}
&$\mathbf{x}^{\prime} \neq \mathbf{x} $ & $\mathbf{z}^{\prime} \neq \mathbf{z}$ & & \\
\hline \hline
\checkmark & & & & 44.59 \\
\checkmark & & &\checkmark &45.08 \\
\checkmark &\checkmark & &\checkmark &45.23 \\
\checkmark & &\checkmark &\checkmark &46.26 \\
\checkmark &\checkmark &\checkmark & &46.61 \\
\checkmark &\checkmark &\checkmark &\checkmark &\textbf{46.95} \\
\hline
\end{tabular}
\caption{Ablation study on Chinese-English translation. $\checkmark$ means that it is included in training.}
\label{table:ablation_study}
\end{table}

\begin{table}[!t]
\centering
\begin{tabular}{l||l|l|l|l}
\hline
\backslashbox[13mm]{${\scriptstyle \gamma_{src}}$}{${\scriptstyle\gamma_{trg}}$} &0.00 &0.25 &0.50 &0.75\\
\hline \hline
0.00 &44.59 &46.19 &46.26 &46.14\\
\hline
0.25 &45.23 &46.72 &\textbf{46.95} &46.52\\
\hline
0.50 &44.25 &45.34 &45.39 &45.94 \\
\hline
0.75 &44.18 &44.98 &45.35 &45.37 \\
\hline
\end{tabular}
\caption{Effect of the ratio value $\gamma_{src}$ and $\gamma_{trg}$ on Chinese-English Translation.}
\label{table:effect_ratio}
\end{table}



\subsection{Ablation Studies}
Table \ref{table:ablation_study} shows the importance of different components in our approach, which include $\mathcal{L}_{clean}$, $\mathcal{L}_{robust}$ and $\mathcal{L}_{lm}$. As for $\mathcal{L}_{robust}$, it includes the source adversarial input, \ie{$\mathbf{x}^{\prime} \neq \mathbf{x}$} and the target source adversarial input, \ie{$\mathbf{z}^{\prime} \neq \mathbf{z}$}. 
In the fourth row with ${\mathbf{x}^{\prime} = \mathbf{x}}$ and ${\mathbf{z}^{\prime} \neq \mathbf{z}}$,  we randomly choose replacement positions of $\mathbf{z}$ since no changes in $\mathbf{x}$ leads not to form the distribution in Eq. ($\ref{eq:trg_pos}$).
We can find removing any component leads to a notable decrease in BLEU. Among those, the adversarial target input ($\mathbf{z}^{\prime} \neq \mathbf{z}$) shows the greatest decrease of $1.87$ BLEU points, and removing language models have the least impact on the BLEU score. However, language models are still important in reducing the size of the candidate set, regularizing word embeddings and generating fluent sentences.

The hyper-parameters $\gamma_{src}$ and $\gamma_{trg}$ control the ratio of word replacement in the source and target inputs. Table \ref{table:effect_ratio} shows their sensitive study result where the row corresponds to $\gamma_{src}$ and the column is $\gamma_{trg}$.
As we see, the performance is relatively insensitive to the values of these hyper-parameters, and the best configuration on the Chinese-English validation set is obtained at $\gamma_{src}=0.25$ and $\gamma_{trg}=0.50$.
We found that a non-zero $\gamma_{trg}$ always yields improvements when compared to the result of $\gamma_{trg}=0$. While $\gamma_{src} = 0.25$ increases BLEU scores for all the values of $\gamma_{trg}$, a larger $\gamma_{src}$ seems to be damaging.

\section{Related Work}
\noindent\textbf{Robust Neural Machine Translation}
Improving robustness has been receiving increasing attention in NMT. For example, \newcite{Belinkov:17,Liu:18,Karpukhin:19,Sperber:17} focused on designing effective synthetic and/or natural noise for NMT using black-box methods. \newcite{Cheng:18} proposed adversarial stability training to improve the robustness on arbitrary noise type. \citet{Ebrahimi:18b} used white-box methods to generate adversarial examples on character-level NMT. Different from prior work, our work uses a white-box method for the word-level NMT model and introduces a new method using doubly adversarial inputs to both attach and defend the model. 

We noticed that \citet{Michel:18} proposed a dataset for testing the machine translation on noisy text. Meanwhile they adopt a domain adaptation method to first train a NMT model on a clean dataset and then finetune it on noisy data. This is different from our setting in which no noisy training data is available. Another difference is that one of our primary goals is to improve NMT models on the standard clean test data. This differs from \citet{Michel:18} whose goal is to improve models on noisy test data. We leave the extension to their setting for future work.


\noindent\textbf{Adversarial Examples Generation}
Our work is inspired by adversarial examples generation, a popular research area in computer vision, \eg in \cite{Szegedy:14,Goodfellow:14b,Moosavi:16}. 
In NLP, many authors endeavored to apply similar ideas to a variety of NLP tasks, such as text classification \cite{Miyato:17,Ebrahimi:18}, machine comprehension \cite{Jia:17}, dialogue generation \cite{Li:17}, machine translation \cite{Belinkov:17}, \etc. Closely related to \cite{Miyato:17} which attacked the text classification models in the embedding space, ours generates adversarial examples based on discrete word replacements. The experiments show that ours achieve better performance on both clean and noisy data.

\noindent\textbf{Data Augmentation}
Our approach can be viewed as a data-augmentation technique using adversarial examples. In fact, incorporating monolingual corpora into NMT has been an important topic \cite{Sennrich:16b, Cheng:16,He:16,Edunov:18}.
There are also papers augmenting a standard dataset based on the parallel corpora by dropping words \cite{Sennrich:16c}, replacing words \cite{Wang:18}, editing rare words \cite{Fadaee:17}, \etc.
Different from these about data-augmentation techniques, our approach is only trained on parallel corpora and outperforms a representative data-augmentation work \cite{Sennrich:16b} trained with extra monolingual data. When monolingual data is included, our approach yields further improvements.

\section{Conclusion}
In this work, we have presented an approach to improving the robustness of the NMT models with doubly adversarial inputs. We have also introduced a white-box method to generate adversarial examples for NMT. Experimental results on Chinese-English and English-German translation tasks demonstrate the capability of our approach to improving both the translation performance and the robustness. In future work, we plan to explore the direction to generate more natural adversarial examples dispensing with word replacements and more advanced defense approaches such as curriculum learning~\cite{jiang2017mentornet,jiang2015self}.

\bibliographystyle{acl_natbib}
\balance
\bibliography{acl2019}
\end{document}